\documentclass[preprint,12pt]{elsarticle}

\usepackage{fullpage}
\usepackage{smile}
\usepackage[colorlinks,
linkcolor=red,
anchorcolor=blue,
citecolor=blue
]{hyperref}
\usepackage{subfigure}
\usepackage{amsmath,amssymb}
\usepackage{enumitem}

\newcommand{\red}{\color{black}}

\newcommand{\mR}{\mathbb{R}}
\newcommand{\mT}{\mathcal{T}}

\newcommand{\hX}{\hat{X}}
\def\mL{\mathcal{L}}
\def\T{{ \mathrm{\scriptscriptstyle T} }}
\def\hA{\hat{A}}
\newcommand{\A}{\mathbb{A}}
\newcommand{\tA}{\tilde{A}}
\newcommand{\tD}{\tilde{D}}
\newcommand{\GCN}{\text{GCN}}
\newcommand{\mZ}{\mathcal{Z}}
\newcommand{\tW}{\tilde{W}}

\journal{Neruocomputing}

\begin{document}

\begin{frontmatter}



\title{AEGCN: An \underline{A}uto\underline{e}ncoder-Constrained \underline{G}raph\\ \underline{C}onvolutional \underline{N}etwork}


\author[1]{Mingyuan Ma}
\author[2]{Sen Na}
\author[1,3]{Hongyu Wang\corref{cor1}}
\ead{why5126@pku.edu.cn}
\address[1]{School of Electronics Engineering and Computer Science, Peking University, Beijing, China}
\address[2]{Department of Statistics, University of Chicago, Chicago, IL, USA}
\address[3]{National Computer Network Emergency Response Technical Team/Coordination Center of China, Beijing, China}
\cortext[cor1]{Corresponding author.}
\begin{abstract}
We propose a novel neural network architecture, called autoencoder-constrained graph convolutional network, to solve node classification task on graph domains. As suggested by its name, the core of this model is a convolutional network operating directly on graphs, whose hidden layers are constrained by an autoencoder. Comparing with vanilla graph convolutional networks, the autoencoder step is added to reduce the information loss brought by Laplacian smoothing. We consider applying our model on both homogeneous graphs and heterogeneous graphs. For homogeneous graphs, the autoencoder approximates to the adjacency matrix of the input graph by taking hidden layer representations as encoder and another one-layer graph convolutional network as decoder. For heterogeneous graphs, since there are multiple adjacency matrices corresponding to different types of edges, the autoencoder approximates to the feature matrix of the input graph instead, and changes the encoder to a particularly designed multi-channel pre-processing network with two layers. In both cases, the error occurred in the autoencoder approximation goes to the penalty term in the loss function. In extensive experiments on citation networks and other heterogeneous graphs, we demonstrate that adding autoencoder constraints significantly improves the performance of graph convolutional networks. Further, we notice that our technique can be applied on graph attention network to improve the performance as well. This reveals the wide applicability of the proposed autoencoder technique.

\end{abstract}



\begin{keyword}
Homogeneous and heterogeneous graphs; Graph convolutional networks; Graph autoencoder; Graph node classification


\end{keyword}

\end{frontmatter}



\section{Introduction}

The complex relationships among data can be represented by networks in a variety of scientific areas, ranging from molecular biology \citep{Mitchell1990Use, Jacobs2001Protein, Higham2008Fitting}, molecular chemistry \citep{Balasubramanian1985Applications, Balaban1985Applications}, genetics \citep{Nordborg2000Linkage, Garroway2008Applications} to sociology \citep{Ugander2011anatomy, Scott2012Social}. Correspondingly, we have protein networks, atom networks, gene networks, and social networks. All these networks data can be structured as graphs. However, in many applications, graphs tend to be high-dimensional and highly entangled. Therefore, how to extract the structural information efficiently is always of central interest.

One of the most notable ways is graph representation learning, which aims to map nodes' high-dimensional representations to low-dimensional vectors and, meanwhile, to preserve the structural information as much as possible. The learned low-dimensional vectors, also called embedding vectors (or embeddings), are capable to benefit a wide range of downstream machine learning tasks, such as link prediction \citep{Taskar2004Link, AlHasan2011survey, Gao2011Temporal}, node classification \citep{Bhagat2011Node, Tang2016Node}, clustering \citep{Tian2014Learning}, community detection \citep{Fortunato2010Community}, and visualization \citep{Maaten2008Visualizing}.

Existing methods of graph representation learning may be categorized into three types:
\begin{enumerate}[label=(\alph*),topsep=0pt]
\setlength\itemsep{0.0em}
\item \textit{random walk-based algorithms}: DeepWalk \citep{Perozzi2014Deepwalk} is regarded as the first widespread embedding method. It interprets the co-occurrence nodes in a random walk as the ``context", applies SkipGram model \citep{Mikolov2013Efficient} on the generated random walks, and maximizes the log-likelihood of observed context nodes. Node2vec \citep{Grover2016node2vec} further extends the idea by proposing a biased random walk algorithm with two hyper-parameters, which balances the exploration-exploitation trade-off and integrates the homophily and structural equivalence of the network into the embeddings. LINE \citep{Tang2015Line} upgrades DeepWalk by defining a novel loss function to preserve both first- and second-order proximity between nodes, and subsequently its extension, PTE \citep{Tang2015Pte}, is developed targeting heterogeneous graphs. See \cite{Dong2017metapath2vec, Fu2017Hin2vec, Wang2017Ice} for more related algorithms and \cite{Cai2018comprehensive} for a brief survey.

\item \textit{matrix factorization-based algorithms}: graph embedding has also been investigated through the lens of matrix factorization. For example, \cite{Tang2009Relational, Tang2011Leveraging} factorize the modularity matrix, which is an effective community measure for many complex networks \citep{Newman2006Modularity}. \cite{Cai2011Graph, Dong2016Learning} factorize the (normalized) Laplacian matrix. \cite{Yang2017Fast, Ou2016Asymmetric} factorize different types of similarity matrices. Moreover, \cite{Qiu2018Network, Liu2019general} provide some theoretical connections between random walk-based algorithms and the spectral theory of graph Laplacian. It is known that DeepWalk is equivalent to implicit matrix factorization on the normalized Laplacian matrix.

\item \textit{graph neural networks (GNNs)}: GNNs are deep learning-based methods that operate directly on graph domains, which have become popular graph analysis methods due to the magnificent performance and the high interpretability. GNNs resolve two limitations of the above two categories: (i) no parameters are shared between nodes so that the problem scale grows linearly with respect to the number of nodes. (ii) the above two categorizes have limited generalization ability so that the learned embeddings perform poorly on new graphs. Motivated by the great success of standard NNs, researchers attempt to generalize NNs to corresponding GNNs. For concreteness, convolutional neural network (CNN) \citep{Krizhevsky2012Imagenet}, recurrent neural network (RNN) \citep{Schuster1997Bidirectional}, and attention network (AN) \citep{Vaswani2017Attention} are generalized to graph convolutional network (GCN) \citep{Kipf2016Semi}, graph recurrent neural network (GRNN) \citep{Li2015Gated, You2018Graphrnn}, and graph attention network (GAT) \citep{Velickovic2017Graph}, respectively. In addition, graph convolutional recurrent network (GCRN) \citep{Seo2018Structured} has also been developed and demonstrated ground-breaking performance on graph data. We point the reader to \cite{Zhou2018Graph} for a recent overview of GNNs. Our paper is closely related to GCN, and more detailed related literature on GCN will be discussed later.

\end{enumerate}

In this manuscript, we complement the third category by a novel GNN architecture, called autoencoder-constrained graph convolutional network (AEGCN). As suggested by its name, our model has two components:
\begin{enumerate}[label=(\alph*),topsep=0pt]
\setlength\itemsep{0.0em}
\item \textit{GCN}: the fundamental thread of our model is GCN, which takes feature matrix and adjacency matrix of a graph as inputs, and outputs node-level representations. These representations are then used in node classification task.

\item \textit{graph autoencoder}: within GCN, we add a graph autoencoder layer to impose some implicit constraints on hidden layers. In particular, the encoder is hidden layer representations while the decoder is one-layer GCN. The autoencoder in our model is used to approximate to either adjacency matrix (for homogeneous case) or feature matrix (for heterogeneous case) of the input graph. It guarantees that the hidden layer does not lose too much \textit{node-level} information from the input, and is equivalent to constraining the hidden layer in a way that the \textit{uniqueness} of each node is encoded properly. The autoencoder approximation error together with the GCN classification error forms the classification objective.

\end{enumerate}
We implement the above model on both homogeneous graphs and heterogeneous graphs for solving node classification task. Comparing with vanilla GCN on homogeneous graphs, we show in experiments that adding an extra autoencoder layer significantly improves the performance. Moreover, we consider heterogeneous graphs that contain different types of nodes and edges. To this end, we design a multi-channel pre-processing network with two layers to compress multiple adjacency matrices to a single adjacency matrix, and apply the autoencoder to approximating to the feature matrix of the input graph. The experimental results back up the argument again that the hidden layer constraints can benefit the performance of GCN. {\red In the heterogeneous case, we also study the reasonability of the autoencoder approximating to the feature matrix instead of to different kinds of adjacency matrices. We realize that approximating to the unique feature matrix has slightly better performance than the best choice of adjacency matrix, which varies with different datasets, and requires much less computations.} Furthermore, we observe in implementations that our autoencoder technique can be effectively utilized on GAT as well, which reveals the wide applicability of our method.

\vskip 4pt
\noindent{\bf Motivation and contribution:} Our model architecture is motivated by recent understanding of GCN. GCN was first proposed in \cite{Kipf2016Semi} for solving semi-supervised node classification problem for graph-structured data. It has then been applied on various application fields due to its simplicity and efficacy \citep{Derr2018Signed, Gao2018Large, Ying2018Graph}. In principle, GCN generalizes the convolution from Euclidean domain to graph domain. It applies Fourier transformation for both signal (or feature) and filter, multiplies them, and transforms the product back to the discrete domain. The transformation relies on the spectral decomposition of graph Laplacian. The low-rank approximation of decomposition is achieved using truncated Chebyshev polynomials. It is known that the benefits of GCN architecture arise from ``local averaging" (or called Laplacian smoothing), brought by the linear approximation of graph convolution. However, there is a trade-off between graph-level information and node-level information.

In specific, Laplacian smoothing nicely integrates the local connectivity patterns by averaging the feature of each node with its neighbors'. The nodes in the same class tend to have a common feature after multi-layer convolutions, and the graph-level information is hence captured. However, a potential concern of such mechanism is \textit{over-smoothing} \citep{Li2018Deeper}. The smoothed features of nodes cannot reflect their uniqueness in the input graph, so that the node-level information is not thoroughly encoded within GCN, which in turn limits the performance of GCN. Motivated by such limitation, we complement GCN with an additional autoencoder layer. The goal of this layer is to reduce the deviation between the hidden layer representations and the original representations, so that the network can preserve much node-level information.

Autoencoder, consisting of encoder and decoder, is a widely used dimension reduction framework. In encoder step, it maps high-dimensional representations to low-dimensional embeddings, while in decoder step, it solves downstream tasks by particular models with embeddings from the encoder step, defines the proper loss function, and trains parameters in both steps. In the present paper, the downstream task is to preserve the node-level information, which is characterized by the approximation to either adjacency matrix or feature matrix. The encoder is naturally given by the hidden layer of GCN, while the decoder we choose is another layer of GCN. We interpret the approximation error as the regularization term in the loss function. Intuitively, in our model, GCN learns the graph-level information by Laplacian smoothing, while autoencoder provides some implicit constraints to alleviate the loss of node-level information.

The main contributions of the paper are three-fold. First, we propose a novel GCN-based network architecture, AEGCN, and apply it on homogeneous graphs. To the best our knowledge, AEGCN is the first algorithm that restricts the hidden layer in a way to avoid over-smoothing, by the aid of the autoencoder framework. The experimental results on citation networks show that AEGCN outperforms other state-of-the-art GCN-based methods. Second, we adjust GCN and the autoencoder technique to fit in the heterogeneous case. We design a multi-channel pre-processing network and, relying on the weighted matrix product of different types of adjacency matrices, obtain a single adjacency matrix, each element of which corresponds to a length-2 meta-path between two nodes. We show in experiments that our model achieves the best performance on multiple heterogeneous graph datasets. Third, we apply the same idea on graph attention network. We observe that the constrained graph attention network also outperforms the original one. This demonstrates the considerable potential of our technique.

\vskip 4pt

\noindent{\bf Relationship to literature:} Our work is related to three active research lines. First, we contribute to the growing literature on GNNs. Like other deep learning models, GNNs achieve very promising results on different tasks and have strong generalization ability. Recent developments on optimization techniques and parallel computation have enabled efficient training on them. As one of the first widespread GNN models, a large body of literature have studied GCN and developed different extensions. For example, \cite{Li2018Adaptive} designs an adaptive GCN model, which is able to take graphs with arbitrary size and connectivity pattern as inputs. \cite{AbuElHaija2018N} aggregates GCN by the information contained in random walks, and generates and trains multiple GCN instances over node pairs for node classification. \cite{Zhao2019T} proposes a temporal GCN model by combining GCN with gated recurrent unit, and applies the model on intelligent traffic systems. In order to improve training efficiency, \cite{Chen2018Fastgcn} interprets the graph convolution as the integral of the embedding function under certain probability measure, and uses the importance sampling to estimate the integral. \cite{Wu2019Simplifying} reduces the complexity of GCN by removing nonlinearities in hidden layers, and illustrates on multiple datasets that such reduction does not negatively affect the accuracy. \cite{He2020LightGCN} also reports LightGCN model to simplify GCN, which only includes the neighborhood aggregation part. Our paper complements the aforementioned GCN-based models with AEGCN, which is the first GCN architecture regularized by autoencoder. Instead of simplifying GCN, we aim to address the over-smoothing issue of GCN via the autoencoder technique.

Second, our work is related to graph autoencoder. Recently, there has been much interest on studying the framework of autoencoder for graph embedding. \cite{Kipf2016Variational} designs an unsupervised autoencoder framework for graphs with a GCN encoder and an inner product decoder. \cite{Berg2017Graph} designs an autoencoder to handle link prediction of the bipartite interaction graphs, and applies it on solving matrix completion problems. \cite{Wang2017Mgae} proposes a marginal graph self-encoder algorithm for graph clustering problem. \cite{Pan2018Adversarially} encodes the topology and node content of the graph into a compact representation, and proposes adversarial graph autoencoder framework via the adversarial training scheme. {\red \cite{park2019symmetric} proposes a symmetric graph autoencoder where the encoder is one-layer GCN and the decoder is designed for ``Laplacian sharpening". Their decoder model intuitively mitigates the effect of Laplacian smoothing in the encoder model, while it also introduces numerical instability into the framework and brings additional challenges in training process.} \cite{Na2020Semiparametric} formalizes the embedding problem as a statistical estimation problem, proposes a semiparametric decoder model, and uses a pseudo-likelihood objective to solve the embedding problem for bipartite graphs. We refer to \cite{Hamilton2017Representation} for a survey on graph autoencoder. Comparing with the above work, our paper heuristically exploits autoencoder on constraining the hidden layers of GCN. The goal of autoencoder in our model is not node classification, which is addressed by GCN, but the approximation to node-level information of the input graph. This implementation of autoencoder has not been considered in the previous work.

Third, our work is related to the literature on understanding the mechanism of GCN. \cite{Li2018Deeper} argues that GCN model is actually a special form of Laplacian smoothing, which is the key reason why GCN works, while in turn results in a concern of over-smoothing. \cite{Kampffmeyer2019Rethinking} studies the trade-off between breadth and depth of GCNs, and adds multiple edges on the original tree-structured data to make it dense to balance the trade-off. However, their method significantly enhances the training difficulty due to the increase of the problem scale. {\red \cite{abu2020n} combines GCN with random walk graph embeddings to train multiple instances of GCNs at different locations on random walks, and learns a combination of the instance outputs. \cite{li2019deepgcns} adapts the techniques from CNN to GCN architectures, including residual/dense connections and dilated convolutions, which are helpful when training a deep GCN. \cite{xu2018representation} explores jumping knowledge networks which leverage different neighborhood ranges to enable better structure-aware representation. \cite{huang2019residual} uses recurrent units to capture the long-term dependency across GNN layers and to identify important information during recursive neighborhood expansion.} All aforementioned works try to improve the performance of GCN by twisting its structure and/or replacing with a more complex architecture. We contribute this series of work by proposing a cheap solution to the potential over-smoothing issue in GCN. We argue that combining autoencoder regularization with GCN is able to effectively avoid over-smoothing of GCN, and such technique is also worth trying for other network architectures.

Throughout the presentation, we let $|V|$ be the cardinality of the set $V$. By $I_n$ we denote the $n\times n$ identity matrix. For a sequence of matrices $A_i\in \mR^{n\times d_i}$, $|\;|_{i}A_i\in \mR^{n\times \sum_{i} d_i}$ denotes the matrix obtained by concatenating $A_i$ by column sequentially.

\vskip 4pt
\noindent{\bf Structure of the paper:} In Section \ref{sec:2}, we introduce some preliminaries including homogeneous and heterogeneous graphs, GCN and autoencoder. In Section \ref{sec:3}, we introduce our model. The experimental results are demonstrated in Section \ref{sec:4} and conclusions and future work are summarized in Section \ref{sec:5}.

\section{Preliminaries}\label{sec:2}

Before introducing our method, we begin with some preliminaries. We first introduce homogeneous graphs and heterogeneous graphs, then present how GCNs implement on homogeneous graphs, and then introduce the autoencoder framework.

Let $G = (V, E)$ be a graph where $V = \{v_1,\ldots, v_n\}$ is a set of nodes and $E$ is a set of edges. We say it is a homogeneous graph if it has a single type of node and a single type of edge, otherwise it is a heterogeneous graph. The former definition is as follows.

\begin{definition}

Given a graph $G = (V, E)$, we let $f^V : V\rightarrow \mT^V$ be a node type mapping function and $f^E: E\rightarrow \mT^E$ be an edge type mapping function, where $\mT^V$ and $\mT^E$ are the predefined nonempty node types set and edge types set. $G$ is called a homogeneous graph if $|\mT^V|+ |\mT^E| = 2$, a heterogeneous graph if $|\mT^V|+ |\mT^E| > 2$.

\end{definition}

\begin{figure}[!htp]
\centering	
\subfigure[Homogeneous graph]	{\label{HoGraph}\includegraphics[width=50mm]{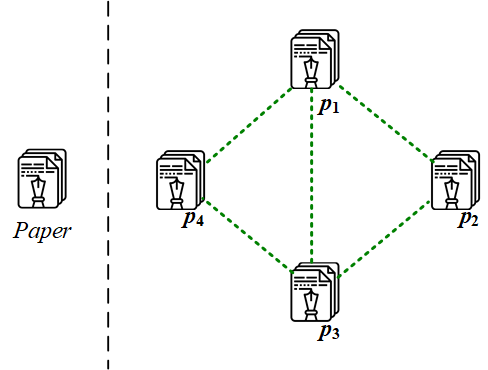}}
\subfigure[Heterogeneous graph]
{\label{HeGraph}\includegraphics[width=70mm]{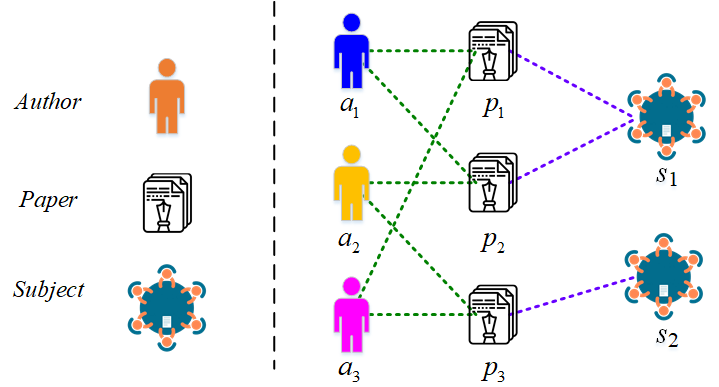}}	
\caption{Illustrative examples of homogeneous graphs and heterogeneous graphs. The left part of the vertical dash line indicates the node type, while the right part is the graph.}\label{fig:1}
\end{figure}

In the above definition, we implicitly assume there exists at least one edge on the graph. In general, heterogeneous graphs \citep{Sun2011Pathsim, Sun2013Pathselclus} contain more comprehensive structured relations (edges) among nodes and unstructured contents (features) associated with each node. For example, features of different types of nodes may fall in different spaces, and two nodes may be connected via different semantic paths, called meta-paths. We demonstrate the differences of two types of graphs by the example in Figure \ref{fig:1}.

In Figure \ref{HoGraph}, we construct a homogeneous graph to model Cora citation network \citep{sen2008collective}. It contains a single node type (paper) and a single edge type (reference relationship). In Figure \ref{HeGraph}, we construct a heterogeneous graph to model ACM citation network. It contains three node types: author, paper, and subject, and four edge types: author-paper, paper-author, paper-subject and subject-paper. For this graph, we see that there are two meta-paths between two paper nodes: paper-author-paper and paper-subject-paper. The former indicates two papers have the same author, while the latter indicates two papers belong to the same subject. Such information can not be represented in a homogeneous graph.

For clarity, we further introduce the notations of adjacency matrix and degree matrix.

\begin{definition}

Given a graph $G = (V, E)$ with $|V| = n$ and $\mT^E$ being the edge types set, we let $\A = \{A_k\}_{k = 1}^{|\mT^E|}$ be the adjacency matrices class, where $A_k \in \mR^{n\times n}$ with $(A_k)_{ij} = 1$ if there exists a type $k$ edge between node $i$ and node $j$, otherwise $(A_k)_{ij}= 0$. When $|\mT^E| = 1$, $\A$ has a single adjacency matrix, denoted by $A$. Moreover, for an adjacency matrix $A$, $\tA = A + I_n$ is the adjacency matrix with self-connections, and the diagonal matrix $\tD \in \mR^{n\times n}$, with $\tD_{ii} = \sum_{j=1}^{n}\tA_{ij}$, is the degree matrix of $\tA$.

\end{definition}

We now set the stage for introducing the graph convolutional network (GCN) on homogeneous graphs. Suppose $A\in\mR^{n\times n}$ and $X \in \mR^{n\times d}$ are the adjacency matrix and the feature matrix of the graph $G$, respectively, GCN has the following propagation rule:
\begin{equation}\label{equ:GCN}
\begin{aligned}
H^{(l+1)}=& \sigma\rbr{\tD^{-\frac{1}{2}} \tA \tD^{-\frac{1}{2}} H^{(l)} W^{(l)}},\quad  l = 0, 1\ldots,\\
H^{(0)} = & X,
\end{aligned}
\end{equation}
where $\sigma(\cdot)$ is some nonlinear activation functions, $H^{(l)} \in \mR^{n\times d_l}$ is the input activation matrix of the $l$-th hidden layer, $W^{(l)}\in \mR^{d_l\times d_{l+1}}$ is the trainable weight matrix with $d_0 = d$. Here, each row of $H^{(l)}$ corresponds to the representation of each node in the hidden layer. The matrix $\tD^{-\frac{1}{2}} \tA \tD^{-\frac{1}{2}}$ comes from applying normalization trick on the convolution matrix $I_n + D^{-\frac{1}{2}}AD^{-\frac{1}{2}}$, where $D$ with $D_{ii} = \sum_{j=1}^nA_{ij}$ is the degree matrix of $A$. The motivation of the rule in \eqref{equ:GCN} is referred to \cite{Kipf2016Semi, Li2018Deeper}. One can show that the convolution on $H^{(l)}$, $\tD^{-\frac{1}{2}} \tA \tD^{-\frac{1}{2}} H^{(l)}$, is equivalent to performing the Laplacian smoothing on each channel of $H^{(l)}$ \citep{Taubin1995signal}, so that the deviation among representations of $H^{(l)}$ will be moderated in $H^{(l+1)}$.

For notation simplicity, we use $\text{GCN}(X, A)$ to denote the output of a GCN model with inputs $X$ and $A$. The number of layers and the choice of activation functions are suppressed in the notation. The standard (non-probabilistic) graph autoencoder model (GAE) \citep{Kipf2016Variational} consists of a GCN encoder model and a nonlinear inner product decoder model. The task of decoder is to reconstruct adjacency matrix $A$. In particular, the autoencoder can be summarized as
\begin{equation}\label{equ:autoencodedr}
\begin{aligned}
&\text{Encoder:} \quad Z = \GCN(X, A),\\
&\text{Decoder:} \quad \hA = \sigma(ZZ^\T).
\end{aligned}
\end{equation}
{\red
It is also worth mentioning that a symmetric autoencoder called GALA is proposed in \cite{park2019symmetric}. The encoder is the same while the decoder is intuitively the inverse of GCN given by
\begin{equation}\label{equ:GALA}
\begin{aligned}
&\text{Encoder:} \quad Z = \GCN(X, A),\\
&\text{Decoder:} \quad \hat{X} = \sigma\rbr{\hat{D}^{-\frac{1}{2}} \bar{A} \hat{D}^{-\frac{1}{2}} Z W}.
\end{aligned}
\end{equation}
where $\bar{A}=2I_n-A$ and $\hat{D}=2I_n+D$. The authors call the decoder Laplacian sharpening. Different from most existing homogeneous autoencoder frameworks where the decoder is applied to reconstruct the adjacency matrix, their decoder reconstructs the feature matrix instead. However, GALA is particularly designed for node clustering and link prediction tasks. More importantly, the symmetry of the model brings numerical instability and requires more involved training process. The experimental results in Section \ref{section:HG} also show that it may not be suitable to be applied on our problems. On the other hand, we are inspired by the great success of GCGA \citep{yu2019real} and GCAE \citep{yan2020graph} models on the fields of real-time traffic speed estimation and building shape recognition, where authors all use another layer of GCN as the decoder. Thus, our method will also replace the decoder in \eqref{equ:autoencodedr} and \eqref{equ:GALA} by GCN.
}

\section{Autoencoder-constrained GCN}\label{sec:3}

This section presents our model. We handle the homogeneous graph first to warm up. The core of our model is GCN, which is used to perform node classification task. We use a two-layer GCN architecture to illustrate the main idea.

Given a homogeneous graph $G = (V, E)$, we suppose $A\in\mR^{n\times n}$ is its adjacency matrix, $X \in \mR^{n\times d}$ is its feature matrix, and $Y \in \mR^{n\times f}$, defined as $Y_{ij} = 1$ if node $i$ belongs to class $j$ and $Y_{ij} = 0$ otherwise, is the label matrix. Following the rule in \eqref{equ:GCN}, two-layer GCN takes the form
\begin{align*}
H^{(1)} = & ReLU\rbr{\tD^{-\frac{1}{2}}\tA\tD^{-\frac{1}{2}} XW^{(0)}},\\
H^{(2)} = & softmax\rbr{\tD^{-\frac{1}{2}}\tA\tD^{-\frac{1}{2}} H^{(1)}W^{(1)}}.
\end{align*}
Here, $W^{(0)}\in \mR^{d\times d_1}$ and $W^{(1)}\in \mR^{d_1\times f}$ are weight matrices for two layers, $ReLU(\cdot) = \max(0, \cdot)$ is applied entry-wise, the softmax activation function, defined as $\rbr{softmax(Z)}_i = \frac{1}{\mZ}\exp(Z_i)$ with $\mZ = \sum_{i}\exp(Z_i)$, is applied row-wise. Based on the output $H^{(2)}$, the classification error is defined by the cross-entropy loss:
\begin{align}\label{equ:loss:class}
\mL_{class} = - \frac{1}{n}\sum_{i=1}^{n}\sum_{j=1}^{f}Y_{ij}\log H^{(2)}_{ij}.
\end{align}

\begin{figure}[!htp]
\centering	
\includegraphics[width=150mm]{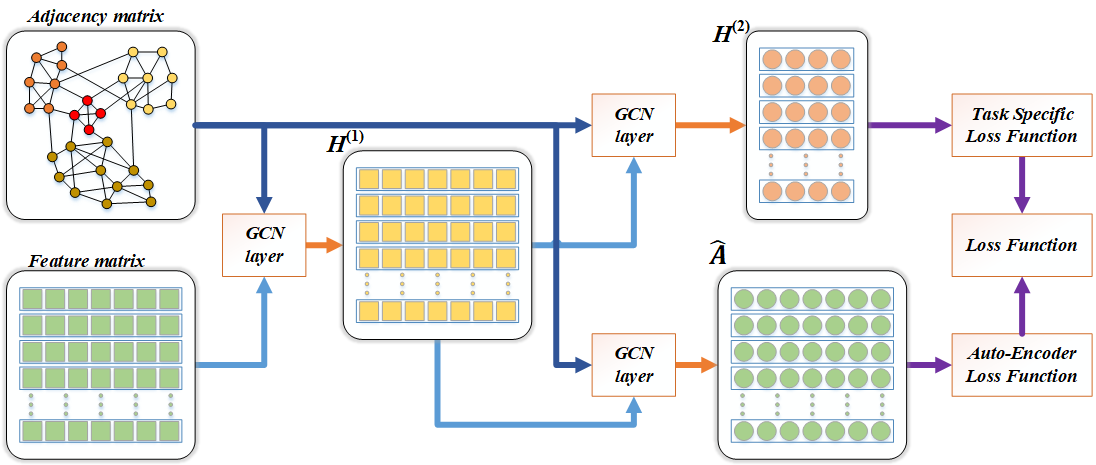}
\caption{Illustration of AEGCN on homogeneous graphs. The top thread is two-layer GCN, while the bottom thread is autoencoder.}\label{fig:2}
\end{figure}

On the other hand, since hidden layer representation $H^{(1)}$ is obtained by doing Laplacian smoothing on $X$, the node-level information of $X$ is depressed in this step. We adopt the autoencoder framework to compensate for such information loss. The encoder model is simply given by $H^{(1)}$, while different from \eqref{equ:autoencodedr} and \eqref{equ:GALA}, the decoder model is another layer of GCN. We have
\begin{align*}
\hA = sigmoid\rbr{ \tD^{-\frac{1}{2}}\tA\tD^{-\frac{1}{2}}H^{(1)}W^{(a)} }.
\end{align*}
Here $W^{(a)} \in \mR^{d_1\times n}$ and sigmoid function is applied entry-wise. We adopt sigmoid instead of softmax as activation function since comparing the output of the decoder with the (normalized) adjacency matrix can be viewed as a multi-label classification task. We realize in implementation that the above decoder model performs better than \eqref{equ:autoencodedr} and \eqref{equ:GALA} on GCN architectures. We then measure the autoencoder approximation error by cross-entropy loss again:
\begin{align}\label{equ:loss:auto}
\mL_{auto} =  - \frac{1}{n^2}\sum_{i,j=1}^{n}(\tD^{-\frac{1}{2}}\tA\tD^{-\frac{1}{2}})_{ij}\log \hA_{ij}.
\end{align}
Combining two loss functions in \eqref{equ:loss:class} and \eqref{equ:loss:auto}, we train $W^{(0)}, W^{(1)}, W^{(a)}$ by performing gradient descent on the penalized loss function
\begin{align*}
\mL = \mL_{class} + \gamma \mL_{auto}
\end{align*}
with tuning parameter $\gamma$. The above network architecture is illustrated in Figure \ref{fig:2}. {\red We should mention that the classification task is performed by two-layer GCN in our framework, and the autoencoder is only used for constraining the hidden layer of GCN. Thus, the autoencoder is used in training set only, while after obtaining trained $W^{(0)}, W^{(1)}$, we apply GCN to do classification in test set without the help of autoencoder.}

Following the same flavor, we consider the heterogeneous graphs. For a heterogeneous graph $G = (V, E)$, we let $X\in \mR^{n\times d}$ and $Y\in \mR^{n\times f}$ be the feature matrix and label matrix, respectively, and $\A = \{A_k\}_{k=1}^{|\mT^E|}$ be the adjacency matrices class. The idea is to transform $\A$ to a single adjacency matrix and apply the same model for the homogeneous case. Inspired by \cite{Yun2019Graph}, we transform $\A$ by two graph transformer layers.

In the first layer, we first generate multiple new graphs defined by the convex combination of adjacency matrices in $\A$. Then, we aggregate graphs to a single graph and use its adjacency matrix as our single adjacency matrix. We make use of the weighted matrix product, so that each entry of the single adjacency matrix corresponds to a meta-path. In particular, we generate $C$ pairs of graph structures from $\A$ by
\begin{align*}
Q_1^i = \sum_{k=1}^{|\mT^E|} \rbr{\tW_{1}^{i}}_k\cdot A_k \quad \text{and}\quad Q_2^i = \sum_{k=1}^{|\mT^E|} \rbr{\tW_{2}^{i}}_k\cdot A_k, \quad \text{for } i = 1, \ldots, C,
\end{align*}
where $\tW_{j}^{i} = softmax\rbr{W_{j}^{i}} \in \mR^{|\mT^E|}$ for $j = 1,2$ are coefficients of the convex combination. Here, $\{W_{1}^{i}, W_{2}^{i}\}_{i = 1}^C$ are weights to be optimized. Given a pair $(Q_1^i, Q_2^i)$, $A^i = Q_1^iQ_2^i$ represents the adjacency matrix of length-2 meth-paths. We further let $\tA^{i} = A^i + I_n$ and the single adjacency matrix is defined as
\begin{align}\label{equ:adj:single}
\tA_H = \sum_{i=1}^{C}\tA^i.
\end{align}

In the second layer, we aggregate the features by one convolutional layer:
\begin{align*}
H^{(0)} = ||_{i=1}^CReLU\rbr{\tD_i^{-1}\tA^i X W^{(\text{aggre})}},
\end{align*}
where $\tD_i$ is the degree matrix of $\tA^i$ and $W^{(\text{aggre})}$ is a trainable weight matrix shared across channels.

\begin{figure}[!htp]
\centering	
\includegraphics[width=150mm]{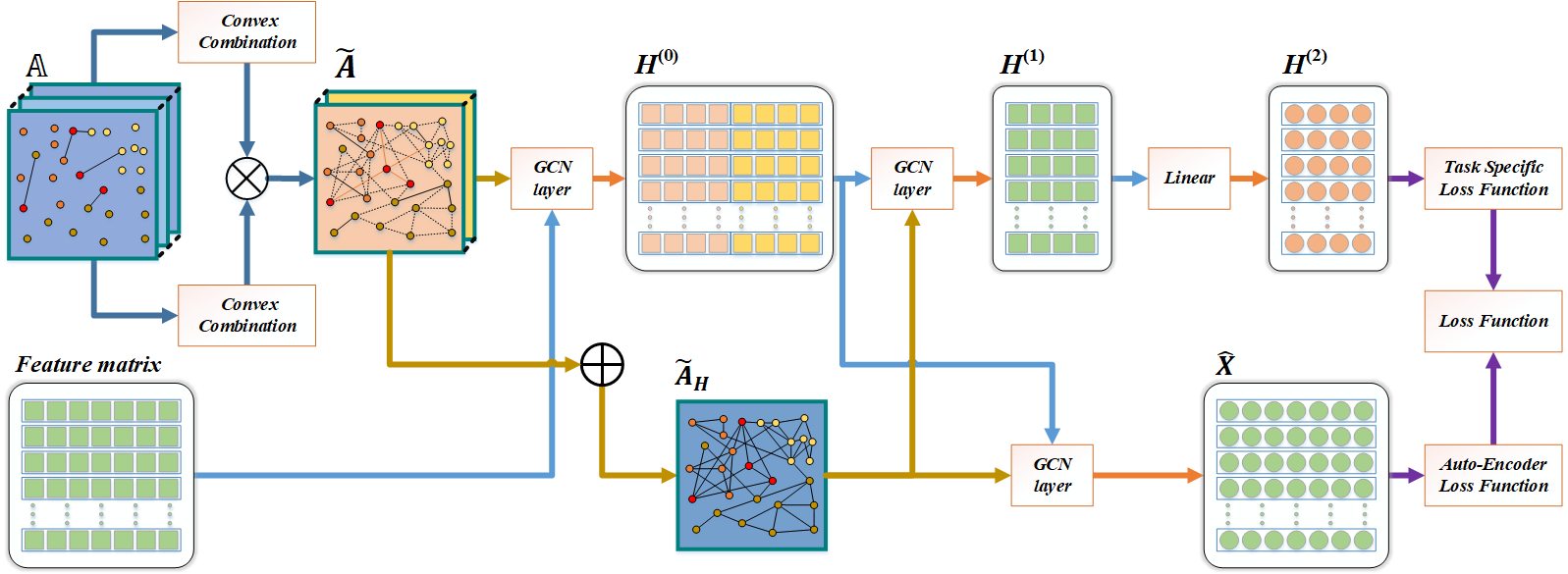}
\caption{Illustration of AEGCN on heterogeneous graphs. The top thread is two-layer GCN, while the bottom thread is autoencoder. The right part from $H^{(0)}$ is consistent with Figure~\ref{fig:2}, while the left part is two graph transformer layers.}\label{fig:3}
\end{figure}

Using $\tA_H$ and $H^{(0)}$ and letting $\tD_H$ be the degree matrix of $\tA_H$, we follow the previous AEGCN architecture:
\begin{align}
\text{GCN}:& \quad H^{(1)} =  ReLU\rbr{\tD_H^{-1}\tA_HH^{(0)}W^{(0)}}, \nonumber\\
&\quad H^{(2)} =  softmax\rbr{H^{(1)}W^{(1)} + b}, \nonumber\\
\text{Autoencoder}: & \quad \hX = sigmoid\rbr{\tD_H^{-1}\tA_HH^{(0)}W^{(a)}}. \label{equ:Auto}
\end{align}
Moreover, the classification error is same as \eqref{equ:loss:class}, while the autoencoder approximation error is redefined as
\begin{align*}
\mL_{auto} = - \frac{1}{nd}\sum_{i=1}^{n}\sum_{j=1}^{d}X_{ij}\log \hX_{ij}.
\end{align*}
Here we keep using cross-entropy loss, since in experiments the input feature matrix $X$ also has $0-1$ elements. Combining loss functions $\mL_{class}$ and $\mL_{auto}$, we perform gradient descent to optimize all weight matrices $\{W_{1}^{i}, W_{2}^{i}\}_{i=1}^C$, $W^{(\text{aggre})}, W^{(0)}, W^{(1)}, W^{(a)}$. We illustrate the heterogeneous case in Figure \ref{fig:3}.

{\red
We should point out that we do not consider a deep GCN with a more complex decoder in both cases. This is for emphasizing the main contribution---the autoencoder regularization---of the paper. The effects of deep decoder regularization on deep GCN have to be further explored. In Section \ref{section:HeG}, we conduct an experiment on a two-layer decoder to initiate such extension and inspire more research in-depth.}

\begin{remark}

There are two main differences between models in heterogeneous case and in homogeneous case.
\begin{enumerate}[label=(\alph*),topsep=0pt]
\setlength\itemsep{0.0em}
\item In heterogeneous case, our autoencoder approximates to the original feature matrix $X$ instead of the adjacency matrix. The reason is that $\A$ contains different types of adjacency matrices. {\red It's not clear to us which adjacency matrix, including $\tA_H$, $A_{all}=\sum{A_k}$ and the entire class $\A = \{A_k\}_{k=1}^{|\mT^E|}$, can reflect the most node-level information of the input graph. Meanwhile, when $n\gg d$ (which is usually the case for heterogeneous graphs), reconstructing the feature matrix $X\in \mR^{n\times d}$ consumes less memory and computation time than reconstructing the adjacency matrix. Thus, we believe the proposed autoencoder is more scalable, though in homogeneous case it is conventional to reconstruct the adjacency matrix \citep{Kipf2016Variational,hasanzadeh2019semi,ke2020graph,ding2020variational}.

\hskip 0.5cm To fully probe the difference, we modify the model and conduct experiments to compare the performance between feature matrix and different adjacency matrices. For $\tA_H$ and $A_{all}$, we simply use the output of \eqref{equ:Auto} to approximate them but here $W^{(a)} \in \mR^{d_0\times n}$, where $d_0$ is the number of columns of $H^{(0)}$. For the entire class $\A$, we let $\{W^{(a, k)}\}_{k = 1}^{|\mT^E|}$ with $W^{(a, k)}\in \mR^{d_0\times n}$ be a sequence of weight matrices and replace $W^{(a)}$ by $W^{(a, k)}$ in \eqref{equ:Auto} to reconstruct $A_k$. This is equivalent to using the weight $W^{(a)} = ||_{k=1}^{|\mT^E|}W^{(a, k)} \in \mR^{d_0\times n|\mT^E|}$ to reconstruct $||_{k=1}^{|\mT^E|}A_k$. The autoencoder loss is cross-entropy loss in \eqref{equ:loss:auto}. The four versions of AEGCN based on $X, \tA_H, A_{all}$, and $\A$ are represented as AEG($X$), AEG($H$), AEG($A$) and AEG($S$) in Section \ref{section:HeG}, respectively. The results in Table \ref{tab:6} show that all four versions of AEGCN achieve better performance than baseline algorithms, and AEG($X$) is slightly better than the best of the other three versions, which varies with different datasets.

}

\item In graph convolution, we use the asymmetric in-degree matrix $\tD_H^{-1}$ to normalize the adjacency matrix, rather than $\tD_H^{-1/2}$ in homogeneous case. This is because most of heterogeneous graphs (e.g. ACM networks and IMDB networks) are directed and symmetric normalization is not suitable.
\end{enumerate}

\end{remark}

\section{Experiments}\label{sec:4}

In this section, we conduct extensive experiments on real benchmarks to testify the proposed method. We also extend our autoencoder regularization idea to graph attention network (GAT) and study the applicability of such technique on different network architectures. Our code is publicly available at \url{https://github.com/AI-luyuan/aegcn}.

\subsection{Homogeneous graphs}\label{section:HG}

We conduct experiments on three commonly used citation networks: Cora, Citeseer, and Pubmed \citep{sen2008collective}. These networks contain only one node type (paper) and one edge type (reference relationship), thus they are homogeneous graphs. The statistics are summarized in Table \ref{tab:1}.

\begin{table}[htp]
\centering
\caption{Statistics of homogeneous graphs}
\label{tab:1}
\setlength{\tabcolsep}{2.2mm}{
	\begin{tabular}{c c c c c c c c}
			\hline
			Dataset & \#Nodes  & \#Edges & \#Classes  & \#Features & \#Training & \#Validation & \#Test \\
			\hline
			Cora & 2708 & 5429 & 7 & 1433 & 140 & 500 & 1000\\
			Citeseer & 3327 & 4732 & 6 & 3703 & 120 & 500 & 1000  \\
			Pubmed & 19717 & 44338 & 3 & 500 & 60 & 500 & 1000\\
			\hline
	\end{tabular}}
\end{table}
{\red
Before introducing the baselines, we test our regularization framework with GALA-based and GCN-based decoders respectively. The results on three datasets are 80.1\% (Cora), 71.1\% (Citeseer), 79.1\% (Pubmed) for GALA-based decoder while 82.4\% (Cora), 72.3\% (Citeseer), 79.3\% (Pubmed) for GCN-based decoder. This shows that Laplacian sharpening does not improve the performance in our node classification problem. In fact, GALA \citep{park2019symmetric} is particularly designed for unsupervised learning. The symmetry of it may bring some advantages in clustering and link prediction tasks, while it causes the training process to be significantly slower than standard GCN-based decoder especially on large datasets. Thus, our method AEGCN sticks on GCN-based decoder instead.
}

We compare our method with several state-of-the-art baselines listed in Table \ref{tab:2}, including some GCN-based methods, DeepWalk, and a semi-supervised embedding method.
\begin{itemize}[topsep=0pt]
\setlength\itemsep{-0.1pt}
\item SemiEmb: A nonlinear semi-supervised embedding algorithm that applies ``shallow"  learning techniques such as kernel methods on deep network architectures, via adding a regularizer at the output layer.

\item DeepWalk: A random walk-based embedding algorithm that applies SkipGram model on the generated random walks and maximizes the log-likelihood objective.

\item GCN: A semi-supervised GNN that generalizes the convolutional network on graph-structured data.

\item FastGCN: A GCN-based method that treats the graph convolution as an integral of embedding functions under certain probability measures, and applies importance sampling to estimate the integral.

\item SGCN: A simplified GCN architecture by  successively removing nonlinearities and collapsing weight matrices between consecutive layers.

\item RGCN: A robust GCN method that is able to absorb the effects of adversarial attacks on the graph by assuming nodes representations of hidden layers follow Gaussian distribution.
\end{itemize}

\begin{table}[htp]
	\centering
	\caption{Descriptions of baseline models for homogeneous graphs}
	\label{tab:2}
	\setlength{\tabcolsep}{7mm}{
		\begin{tabular}{l|l}
			\hline
			Model  & Short description \\
			\hline
			\hline
			SemiEmb \citep{Weston2012Deep}&  Deep learning via semi-supervised embedding  \\
			\hline
			DeepWalk \citep{Perozzi2014Deepwalk}& Random walk-based network embedding method \\
            \hline
			GCN \citep{Kipf2016Semi}& Graph convolutional network  \\
			\hline
			FastGCN \citep{Chen2018Fastgcn}& Fast learning with GCN via importance sampling\\		
			\hline
			SGCN \citep{Wu2019Simplifying}& Simplifying graph convolutional network \\		
			\hline
			RGCN \citep{Zhu2019Robust}& Robust graph convolutional network \\		
			\hline
	\end{tabular}}
\end{table}

\vskip 4pt
\noindent{\bf Implementation details:} As described in Section \ref{sec:3}, our model consists of two GCN layers and one autoencoder layer. We use the same data splitting as in \cite{yang2016revisiting} as shown in Table \ref{tab:1}. For all datasets, we train the model for 200 epochs, tune hyperparameters, including hidden layer dimension, learning rate, dropout rate and weight decay, for Cora dataset only, and use the same set of parameters for Citeseer and Pubmed. The hidden layer dimension $d_1$ is set to be 18 and, same as \cite{Kipf2016Semi}, the learning rate, dropout rate, weight decay are set to be 0.01, 0.5, 0.0005, respectively. We set $\gamma = 10$ for Core and Citeseer, and set $\gamma = 0.001$ for Pubmed. {\red We apply random initialization to all learnable parameters.} All results are reported over 30 independent runs. The results for all other baselines are directly borrowed from \cite{Kipf2016Semi, Wu2019Simplifying, Zhu2019Robust}.

\vskip 4pt
\noindent{\bf Experimental results:} The results are summarized in Table \ref{tab:3}. From the table, we find that GNN methods enjoy significantly better performance than SemiEmb and DeepWalk. Comparing within GNNs, AEGCN has consistently better results on all three datasets than GCN, FastGCN, and SGCN. RGCN performs slightly better than AEGCN on Cora network, while in turn AEGCN outperforms than it on the other two networks. By simple calculations, we see that our simple autoencoder constraint framework can improve the accuracy of other GCN-based methods by 0.2\% to 3.5\%. This suggests the effectiveness of the introduced autoencoder constraints. As for efficacy, we find that AEGCN only adds one GCN layer compared to the vanilla GCN, so they have comparable running time.

\begin{table}[htp]
	\centering
	\caption{Comparison results of homogeneous graphs}
	\label{tab:3}
	\setlength{\tabcolsep}{5mm}{
		\begin{tabular}{llll}
			\hline
			Method & Cora & Citeseer & Pubmed \\
			\hline
			\hline
			SemiEmb   & 59.0 & 59.6 & 71.1   \\
			DeepWalk  & 67.2 & 43.2 & 65.3   \\
			GCN       & 81.5 & 70.3 & 79.0   \\
			FastGCN   & 79.8 $\pm$ 0.3 & 68.8 $\pm$ 0.6 & 77.4 $\pm$ 0.3   \\
			SGCN       & 81.0 $\pm$ 0.0 & 71.9 $\pm$ 0.1 & 78.9 $\pm$ 0.0   \\
			RGCN      & \textbf{82.8 $\pm$ 0.6} & 71.2 $\pm$ 0.5 & 79.1 $\pm$ 0.3   \\
			AEGCN     & 82.4 $\pm$ 0.7 & \textbf{72.3 $\pm$ 0.6} & \textbf{79.3 $\pm$ 0.1}   \\
			\hline
	\end{tabular}}
\end{table}

\subsection{Heterogeneous graphs}\label{section:HeG}

We consider two heterogeneous graphs: ACM citation network and IMDB movie network. ACM contains three types of nodes (paper, author, subject), and four types of edges (paper-author, author-paper, paper-subject, subject-paper). The category of the paper is the label, and the feature of each node is given by bag-of-words of keywords. IMDB contains three types of nodes (movie, actor, director), and four types of edges (movie-actor, actor-movie, movie-director, director-movie). The genre of the movie is the label, and the feature of each node is given by bag-of-words representations of plots. The statistics of two datasets are summarized in Table \ref{tab:4}.

\begin{table}[htp]
	\centering
	\caption{ Statistics of heterogeneous graphs}
	\label{tab:4}
	\setlength{\tabcolsep}{2.0mm}{
		\begin{tabular}{c c c c c c c c}
			\hline
			Dataset & \#Nodes  & \#Edges & \#Edge type  & \#Features & \#Training & \#Validation & \#Test \\
			\hline
			ACM & 8994 & 25922 & 4 & 1902 & 600 & 300&  2125   \\
			IMDB & 12772 & 37288 & 4 & 1256 & 300 & 300 & 2339 \\
			\hline
	\end{tabular}}
\end{table}

In this part, we compare our four versions of AEGCN with four baselines listed in Table \ref{tab:5}. The GCN-based baselines in Table~\ref{tab:2} are not implemented here since they are particularly designed for homogeneous graphs, and it's not clear how to fairly adapt them to heterogeneous graphs.

\begin{itemize}[topsep=0pt]
\setlength\itemsep{-0.1pt}
\item metapath2vec: A random walk-based embedding method for heterogeneous networks, which constructs the heterogeneous neighborhood of a node by performing meta-path-based random walks, and hinges on a heterogeneous SkipGram model
to compute embeddings.

\item HAN: A semi-supervised GNN for heterogeneous network which generates node embedding by aggregating features from meta-path-based neighbors in a hierarchical manner to employ both node-level attention and graph-level attention.

\item GTN: A supervised GNN for heterogeneous network which generates multiple meta-paths by matrix multiplication, applies GCN on each path, and stacks all learned embeddings.

\end{itemize}

\begin{table}[htp]
	\centering
	\caption{Descriptions of baseline models for heterogeneous graphs}
	\label{tab:5}
	\setlength{\tabcolsep}{7mm}{
		\begin{tabular}{l|l}
			%	\begin{tabular}{ |c|c|c|c|c|c|c|c|c|c| }
			\hline
			Model  & Short description \\
			\hline
			\hline
			DeepWalk \citep{Perozzi2014Deepwalk} & Random walk-based network embedding method  \\
			\hline
			metapath2vec \citep{Dong2017metapath2vec} & Random walk-based network embedding method \\
			\hline
			HAN \citep{Wang2019Heterogeneous}& GNN for heterogeneous graph \\		
			\hline
			GTN \citep{Yun2019Graph}& GNN for heterogeneous graph \\		
			\hline
	\end{tabular}}
\end{table}

\vskip4pt

\noindent{\bf Implementation details:} As described in Section \ref{sec:3}, our model consists of two pre-processing graph transformer layers, two GCN layers, and one autoencoder layer. We use the same data splitting as in \cite{Yun2019Graph} as shown in Table \ref{tab:4}. We train the model with a maximum of $40$ epochs for ACM and $20$ epochs for IMDB. For both datasets, the hidden layer dimensions $d_0$ (columns of $H^{(0)}$) and $d_1$ (columns of $H^{(1)}$), and the number of channels $C$ are set to be 128, 64 and 2. The learning rate, weight decay, and regularization parameter $\gamma$ are set to be 0.005, 0.001, and 1, respectively. Same as GTN, dropout technology is not used for AEGCN in the experiment. Results are reported over 10 independent runs. Results for DeepWalk, metapath2vec and HAN are borrowed from \cite{Yun2019Graph}. Results for GTN are reported using the source code from \cite{Yun2019Graph} under the same parameters setup and same experimental environment.

\vskip 4pt
\noindent{\bf Experimental results:} The results are summarized in Table \ref{tab:6}. From the table, we observe that four versions of AEGCN have the best performance on both datasets. DeepWalk and metapath2vec perform worse than GNN methods. Although HAN is a modified GAT for heterogeneous graph, the GCN-based models, GTN and AEGCN, perform better than HAN. By Table \ref{tab:6}, we reach the same conclusion that the proposed architecture is effective for solving node classification task on heterogeneous graphs. {\red Moreover, within four versions of AEGCN, we see that AEG$(X)$ steadily performs slightly better than the best of the other three versions, which changes depending on datasets. For example, AEG$(H)$ performs the best among the adjacency matrix approximation for ACM, while AEG$(A)$ performs the best for IMDB. However, both of them perform worse than AEG($X$). Further, since both datasets have $d\ll n$, we find that AEG$(X)$ requires much less computations in training.}

\begin{table}[htp]
	\centering
	\caption{Comparison with baseline models}
	\label{tab:6}
	\setlength{\tabcolsep}{1.5mm}{
		\begin{tabular}{c| c c c c c c c c}
			\hline
			Dataset  & DeepWalk & metapath2vec & HAN & GTN & AEG$(H)$ & AEG$(S)$ & AEG$(A)$ & AEG$(X)$\\
			\hline
			\hline
			ACM & 67.42 & 87.61& 90.96 & 92.65& 92.93 & 92.78& 92.68& \textbf{93.08}\\
			IMDB & 32.08 &35.21& 56.77  &  57.29& 59.86 & 59.17& 60.18& \textbf{60.27}\\
			\hline
	\end{tabular}}
\end{table}

\begin{figure}[!htp]
\centering	
\includegraphics[width=150mm]{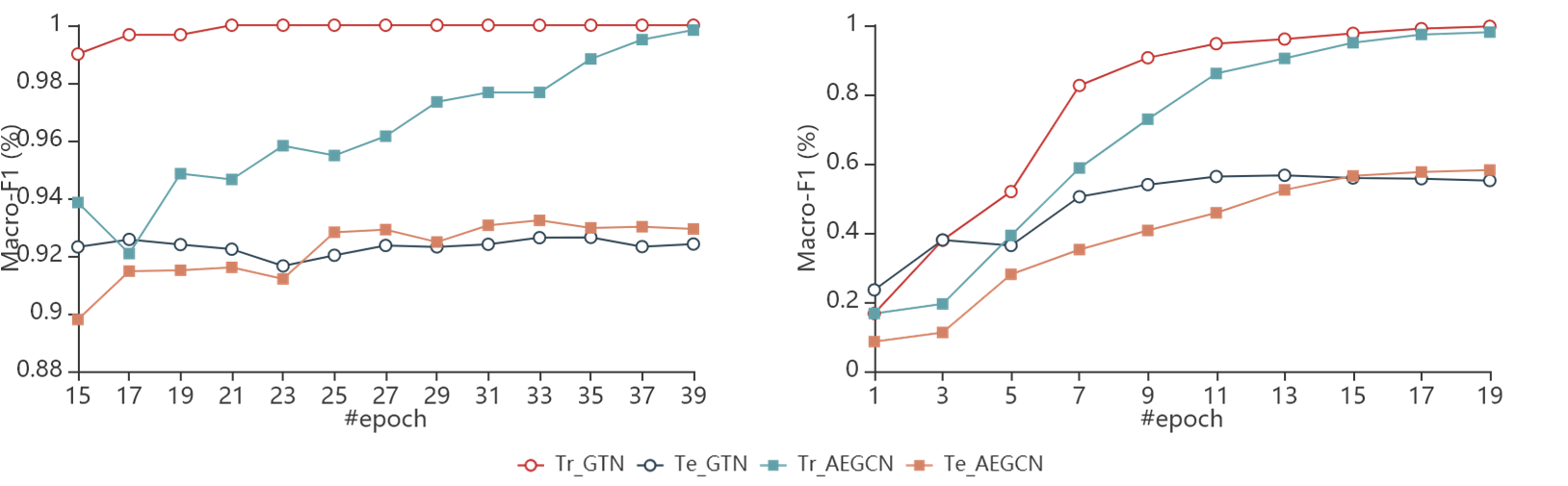}
\caption{Macro-F1 score trajectories of GTN and AEGCN. The left panel is the result on ACM network, while the right panel is the result on IMDB network. For both panels, Tr\_GTN and Tr\_AEGCN correspond to the F1 score on the training set, while Te\_GTN and Te\_AEGCN correspond to the F1 score on the test set.}\label{fig:4}
\end{figure}

To have a closer view, we plot in Figure \ref{fig:4} the changes in Macro-F1 score during AEGCN and GTN training processes in two datasets. Macro-F1 score is a common metric to measure the accuracy of the classifier, which takes both precision and recall into account. In general, the higher the F1 score, the more accurate the classifier. However, one should be aware of an extremely high F1 score on the training set, which is generally due to the overfitting error. From two plots in Figure \ref{fig:4}, we see that F1 score grows gradually during the training process of AEGCN, while it almost hits 1 after very few epochs for GTN. This observation suggests that our proposed regularized method can effectively postpone the occurrence of overfitting during the training. On the other hand, we also note that AEGCN has higher F1 score on the test set finally. Therefore, AEGCN is favorable on both aspects.

{\red
Furthermore, we implement a two-layer GCN decoder to explore the performance of a deep decoder. The results of four versions of AEGCN are summarized in Table \ref{tab:twolayer}, where $*$ represent the methods using two-layer GCN decoder. From Table \ref{tab:twolayer}, we see that deeper decoder generally performs better except for the AEG($H$). Thus, we believe that a deeper autoencoder is helpful in our framework, though a more thorough comparison is deferred to the future work.

\begin{table}[htp]
	\centering
	\caption{Comparison with deep decoder}
	\label{tab:twolayer}
	\setlength{\tabcolsep}{1.0mm}{
		\begin{tabular}{c| c c c c c c c c}
			\hline
			Dataset  & AEG$(H)$ & AEG$(H)^*$ & AEG$(S)$ & AEG$(S)^*$ & AEG$(A)$ & AEG$(A)^*$& AEG$(X)$ & AEG$(X)^*$\\
			\hline
			\hline
			ACM & 92.93& 92.83 & 92.78& 92.88 & 92.65& 92.74 & 93.08& 93.16\\
			IMDB &59.86& 59.71 & 59.17& 61.27 &  60.18& 60.97 & 60.27& 60.38\\
			\hline
			\multicolumn{4}{l}{* refers to a two-layer GCN decoder}
	\end{tabular}}
\end{table}

}

\subsection{Autoencoder on GAT}

We apply our autoencoder regularization idea on GAT to explore the extensibility of our technique. We add the autoencoder layer on the hidden representations before the classification layer of GAT, and conduct experiments on citation networks. We let $\gamma=1$ for all three networks and use the same parameters setting as \cite{Velickovic2017Graph}. {\red For fair comparison, results of GAT are reported using the sparse version of the source code of \cite{Velickovic2017Graph}, which is the same as what the encoder of AEGAT uses.} The results are reported over 10 runs.

\begin{table}[htp]
	\centering
	\caption{Comparison with GAT models}
	\label{tab:7}
	\setlength{\tabcolsep}{5mm}{
		\begin{tabular}{llll}
			\hline
			Method & Citeseer & Cora & Pubmed \\
			\hline
			\hline
			GAT      & 72.1 $\pm$ 0.9          & 83.3 $\pm$ 0.6          & 77.2 $\pm$ 1.0   \\
			AEGAT     & \textbf{72.6 $\pm$ 0.6} & \textbf{83.8 $\pm$ 0.3} & \textbf{78.5 $\pm$ 0.3}   \\
			\hline
	\end{tabular}}
\end{table}

The results are shown in Table \ref{tab:7}. From the table, we see AEGAT performs better on the Cora, Citeseer and Pubmed networks. Throughout the implementation, we simply use the original parameters of GAT with manually specified $\gamma$, and do not tune parameters during the training. We believe that better results are achievable if we further tune parameters for AEGAT sophisticatedly.

This implementation shows that our autoencoder regularization framework not only benefits GCN, but also can migrate to other graph network architectures to improve the performance. This observation highlights the wide applicability and considerable potential of the proposed technique.

\section{Conclusion and future work}\label{sec:5}

In this paper, we propose a novel graph neural network architecture, called autoencoder-constrained graph convolutional network, abbreviated to AEGCN. The core of AEGCN is GCN, which is used to perform node classification task. Within GCN, we impose an autoencoder layer to reduce the loss of node-level information. The error occurred in the autoencoder step is treated as the regularizer of the classification objective. We apply our model on homogeneous graphs and heterogeneous graphs, and achieve superior performance on both cases over other competing baselines. We also apply our idea of autoencoder constraints on graph attention network, and find it can improve the performance of GAT as well. This observation reveals that our technique is applicable for a potentially wide range of network architectures.

Interesting future directions include studying the effectiveness of autoencoder constraints on other types of GNN architectures. In addition, the effects of the autoencoder regularization on deeper GNNs have to be further explored.

\section*{Acknowledgements}

The work is supported by the National Key R\&D Program of China (2019YFA0706401), National Natural Science Foundation of China (61802009, 61902005). Mingyuan Ma is partially supported by Beijing development institute at Peking University through award PkuPhD2019006. Sen Na is supported by Harper Dissertation Fellowship from UChicago.



%%       \citet{<label>} ==> Jones et al. [21]
%%       \citep{<label>} ==> [21]
%%



\bibliographystyle{elsarticle-num-names}
\bibliography{paper}

%\begin{thebibliography}{00}



%\end{thebibliography}

\end{document}